\documentclass[10pt,twocolumn,letterpaper]{article}

\usepackage{iccv}
\usepackage{times}
\usepackage{epsfig}
\usepackage{graphicx}
\usepackage{amsmath}
\usepackage{amssymb}

\usepackage{color}
\usepackage{multirow}
\usepackage{listings}
\usepackage{makecell}

\usepackage[breaklinks=true,bookmarks=false]{hyperref}

\iccvfinalcopy 

\def\iccvPaperID{****} 

\ificcvfinal\fi

\begin{document}

\title{LLA: Loss-aware Label Assignment for Dense Pedestrian Detection}

\author{Zheng Ge$^{1,2}$, Jianfeng Wang$^2$, Xin Huang$^1$, Songtao Liu$^2$, Osamu Yoshie$^1$\\
$^1$Waseda University, $^2$Megvii Technology\\
\tt\small jokerzz@fuji.waseda.jp;wangjianfeng@megvii.com;koushin@toki.waseda.jp;\\ 
 \tt\small liusongtao@megvii.com;yoshie@waseda.jp}

\maketitle
\ificcvfinal\thispagestyle{empty}\fi

\begin{abstract}
Label assignment has been widely studied in general object detection because of its great impact on detectors' performance. However, none of these works focus on label assignment in dense pedestrian detection. In this paper, we propose a simple yet effective assigning strategy called \textbf{L}oss-aware \textbf{L}abel \textbf{A}ssignment (LLA) to boost the performance of pedestrian detectors in crowd scenarios. LLA first calculates classification (cls) and regression (reg) losses between each anchor and ground-truth (GT) pair. A joint loss is then defined as the weighted summation of cls and reg losses as the assigning indicator. Finally, anchors with top $K$ minimum joint losses for a certain GT box are assigned as its positive anchors. Anchors that are not assigned to any GT box are considered negative. Loss-aware label assignment is based on an observation that anchors with lower joint loss usually contain richer semantic information and thus can better represent their corresponding GT boxes. Experiments on CrowdHuman and CityPersons show that such a simple label assigning strategy can boost MR by 9.53\% and 5.47\% on two famous one-stage detectors -- RetinaNet and FCOS, respectively, demonstrating the effectiveness of LLA. The code is available at \url{https://github.com/Megvii-BaseDetection/LLA}. 
\end{abstract}

\section{Introduction}

Pedestrian detection in crowd scenarios has attracted considerable attentions in the recent literature~\cite{wang2018repulsion,citypersons,crowdhuman,adaptivenms} and applications (\emph{e.g.}, autonomous driving and video surveillance). It is widely used in many real-world scenarios where the density of people is high, \emph{i.e.}, airports, train stations, shopping malls \emph{etc}. Compared to general object detection~\cite{coco}, target objects in crowd scenarios are more densely arranged, resulting in heavy occlusion between different objects. Although the performance of modern Convolutional Neural Network (CNN) based object detectors~\cite{retinanet, fcos, faster-rcnn, yolo} is growing rapidly, they usually suffer when applied to dense pedestrian detection (DPD). This is mainly caused by following two issues: mis-classified occluded pedestrian and mis-placed detected results. Recent works mainly utilized additional information or regularization term to relieve these two problems. For example, Bi-box~\cite{bibox} and R2-NMS~\cite{r2nms} alleviate the first issue by introducing visible body annotations as extra supervisions. For the second issue, ~\cite{wang2018repulsion} imposes a novel regression penalty term on the misplacing predictions to tackle it. While these methods try to amend the poor predictions from the detectors, in this paper, we delve into the essential cause of these issues and find an important problem which has never been discussed in the literature of DPD -- label assignment.

Anchor box is the basic processing unit in CNN based object detectors. The procedure in which anchors are assigned as positive or negative during training is called \textit{label assignment}. For anchor-based detectors, the most common label assigning strategy usually utilizes Intersection-over-Union (IoU) between anchors and a ground-truth (GT) bounding boxes. For example, in RetinaNet~\cite{retinanet}, if an anchor's IoU with a certain GT box exceeds 0.5, this anchor is considered as \emph{positive} and assigned to that GT box. Otherwise, anchors are assigned as \emph{negative} or \emph{ignore} according to its maximum IoU between GT boxes. For anchor-free detectors, we first denote the dense sample locations on feature maps as \textbf{anchor points}. In a typical anchor-free detector -- FCOS~\cite{fcos}, a group of anchor points are directly assigned as positive if they fall into a square region in the center of a GT box. Such hand-crafted assigning strategies tend to assign positive anchors near the geometric center of that GT box. While they are proved to work effectively in many famous detectors~\cite{retinanet, fcos, faster-rcnn, yolo}, things are different when they come to pedestrian detection in crowd scenarios. If a person is heavily occluded, his/her geometric center may fall onto other's body, which will lead to inconsistency between the features of sampled points and their corresponding GT boxes. These twisty samples interfere with the training of detectors are certainly one of the main reasons for the mis-classifying and misplacing issues in DPD.

Several recent works~\cite{atss, paa, freeanchor} try to make the procedure of label assignment more adaptive for general object detection. A typical pipeline of them follows: 1) Constructing a bag of positive candidate anchors for each GT. 2) Calculating a certain metric \emph{e.g.} IoU~\cite{atss}, score function~\cite{paa} or likelihood~\cite{freeanchor} for each GT's candidate anchors. 3) Applying statistical tools or hard thresholds on the calculated metric to define positive and negative anchors. Although they can adaptively define what is positive and negative according to the network's prediction, they share a common prior that a set of positive candidate anchors need to be constructed in advance, which will be based on hand-crafted rule -- IoU to ensure that reasonable statistic values can be acquired. Such a constraint limits the positive regions near the geometric center of each GT box. As we stated above, in dense pedestrian detection, the anchor boxes/points near the center of a GT box could be improper and even harmful for being positives if the corresponding person is heavily occluded. These methods fail to take this case into consideration, making their assigning results sub-optimal in crowd scenarios.

To break the limit of current existing label assigning strategies, we propose an extremely simple but effective label assigning strategy called \textbf{L}oss-aware \textbf{L}abel \textbf{A}ssignment (LLA) for dense pedestrian detection. First, LLA calculates \emph{cls} and \emph{reg} losses between each anchor and GT pair. Then, the weighted summation of \emph{cls} and \emph{reg} losses is defined as the joint loss to estimate how well can one anchor learn one GT box. To help model converge better, we impose an ``in box'' term $C^{inbox}$ into the loss term as a minimal constraint. Specifically, if the center of an anchor box fails to fall into any GT box, we will add a constant punishment term $C^{inbox}\ (C^{inbox}>0)$ in its joint loss. Otherwise, $C^{inbox}$ is set to $0$. Finally, anchors with top $K$ minimum joint losses for a certain GT box are assigned as its positive anchors. Anchors that are not assigned to any GT box are considered negative. Noted that in LLA, label assignment fully abandons the scale prior as proposed in FPN~\cite{fpn}, and center prior as utilized in FreeAnchor/FCOS/ATSS. LLA defines positive anchors based on the model's output, making LLA fully adaptive. LLA is proved to be \textbf{occlusion-aware} because heavily occluded regions tend to have a higher joint loss and thus are less likely to be assigned as positive. On CrowdHuman~\cite{crowdhuman}, LLA brings 9.53\% and 5.47\% improvements on MR when applied to RetinaNet and FCOS, respectively, demonstrating its effectiveness. Experiments on CityPersons~\cite{citypersons} further reveals LLA's capability on various pedestrian detection datasets. 

\section{Related Works}
\subsection{Occlusion Handling for Pedestrian Detection.} 
Recently, occlusion handling becomes a popular topic in the field of pedestrian detection. In heavy occlusion situations, the detector will get confused by the adjacent instance which leads to an inaccurate boundaries regression. To solve this problem, Repulsion Loss~\cite{wang2018repulsion} and OR-CNN~\cite{zhang2018occlusion} both impose additional penalty terms on the BBoxes that appear in the middle of two persons to force them to regress to the right person. Moreover, utilizing visible annotation as extra supervision is a common strategy to obtain more precision location information. Bi-box~\cite{bibox} adds a visible branch on Fast R-CNN to predict the full and visible body of a pedestrian at the same time. ATT~\cite{zhang2018occluded} exploits the visible-region information as external guidance to handle various occlusion patterns in crowded situations. MGAN~\cite{pang2019mask} forces the detector to focus on the visible regions of a pedestrian by adopting a novel attention branch to highlight the visible body region while suppressing the occluded part. What's more, some researchers point out that in crowded scenario NMS may be trapped in a dilemma: a lower threshold of intersection over union (IoU) resulting in the miss of highly overlapped pedestrians while a higher IoU threshold naturally brings in more false positives. To solve this problem, Adaptive-NMS~\cite{adaptivenms} proposes a subnet to predict the threshold for different anchors. R2-NMS~\cite{r2nms} leverages the less occluded visible parts to remove the redundant boxes. PS-RCNN~\cite{psrcnn} utilizes two parallel R-CNN modules to detect slightly/none occluded and heavily occluded human instances in a divide-and-conquer manner. Different from all the existing works, our method handles the severe occlusion situation by a novel label assigning method, which neither requires additional annotation nor demands extra parameters. 

\subsection{Hand-crafted Label Assignment}

Anchors are a set of pre-defined square boxes with different scale and aspect ratios which are densely assigned to each spatial location on feature maps. Traditional anchor-based object detectors~\cite{retinanet, faster-rcnn, fpn, cascadercnn} assign labels for anchors based on hand-crafted IoU between anchors and GTs. Specifically, if an anchor's IoU with a certain GT box exceeds a certain threshold (\emph{e.g}. 0.5 for RetinaNet~\cite{retinanet}), this anchor is defined as positive. The remaining anchors are either defined as \textit{negative} or \textit{ignore} according to its maximum IoU between GT boxes. Besides, extra scale constraint introduced by FPN~\cite{fpn} is imposed into these detectors~\cite{fpn, cascadercnn} to better handle the scale variations in objects. Instead of using pre-defined anchors, MetaAnchor~\cite{metaanchor} proposes an anchor function to provide anchors with dynamic shapes in both training and testing phase. Guided Anchoring~\cite{guidedanchor} first learns a set of adjusted anchors in an anchor-free manner to better fit the shape of targets, and then design the subsequent anchor-based modules based on the learned anchors.

Anchor-free methods have drawn more and more attention recently due to its simple pipeline. FCOS~\cite{fcos} and FoveaBox~\cite{foveabox} define anchors in the center region of targets as positive anchors.
FSAF~\cite{fsaf} further utilizes an online Feature Selection Module to select the appropriate feature level for each GT. In keypoint-based anchor-free detectors~\cite{cornernet, extremenet, centernet, reppoints}, only a single center point for each GT (anchor which is the closest to the center of that GT) is defined as positive, while other anchors are all negatives. Because the mechanism of keypoint-based detectors is much different from bounding box based detectors, they are out of the scope of this paper.

\subsection{Dynamic Label Assignment}
Recently, methods called dynamic label assignment are proposed to improve the training process of detectors. ATSS\cite{atss} introduces dynamic IoU thresholds by calculating the mean and standard deviation of IoU values on a set of pre-defined candidate anchors for each GT. FreeAnchor~\cite{freeanchor} designs a detection-customized likelihood that takes precision and recall into consideration to tackle the anchor-object matching problem. Specifically, anchors that have a higher likelihood are defined as positives. PAA~\cite{paa} first designs a score function based on the classification and regression loss then applies one-dimensional GMM~\cite{gmm} for these calculated scores for each GT to choose the thresholds for separating positive and negative anchors. DeTR~\cite{DeTR} and DeFCN~\cite{DeFCN} explore customized loss and quality terms as their indicators for one-to-one label assignment. Although one-to-one matching strategy is proved crucial to end-to-end detectors, it may not be the optimal choice for detectors followed by NMS.

Instead of solving the anchor-object matching problem directly, there are also a few works trying to re-weight the positives and negatives which can be categorized into generalized label assignment. PISA~\cite{pisa} first proposes two ranking strategies -- IoU-HLR and Score-HLR to rank positive and negative proposals, respectively, to evaluate the importance of anchors, then forces the model focus on more important samples by giving them higher weights. Noisy Anchor~\cite{noisyanchor} constructs a cleanliness score based on the detector's outputs to re-weight anchors and soften classification labels. However, as we stated above, although they can adaptively define positive/negative anchors, they still limit the positive regions near the geometric center of each GT box, which could harm the detectors' performance in DPD.

\section{Method}

\subsection{Revisiting Label Assignment in General Object Detection}
Let us take a brief look at how label assignment is conducted on two well-known one-stage detectors -- RetinaNet and FCOS. Given an input image $M$, the ground-truth annotations are denoted as $G$, where a ground-truth box $g_i \in G$ is made up of a class label $g_i^{cls}$ and a location $g_i^{loc}$. Due to the wide-spread of multi-scale feature pyramids network (FPN), both scale and spatial constraints need to be considered during label assignment.

In RetinaNet, $a_j \in A$ stands for an \textbf{anchor box}. RetinaNet handles spatial and scale constraint simultaneously based on the Intersection-over-Union (IoU) matching rule. During training, $a_j$ is assigned to GT $g_i$ if $IoU(a_j, g_i^{loc})>0.5$, while $a_j$ is defined as negative if $ {\forall} g_i \in G, IoU(a_j, b_i^{loc})<0.4$. Anchors which are neither positive nor negative are ignored at that training step. 

In FCOS, $a_j \in A$ stands for an \textbf{anchor point}. During training, $a_j$ is assigned to GT $g_i$ only if 1)$a_j$ falls into the center area (within a fixed radius) of $g_i$. 2) $a_j$ meets the pre-defined scale constraint introduced by FPN\cite{fpn}. Anchor points which do not meet these two requirements are defined as negatives. Note that in FCOS, both spatial and scale constraints are explicitly imposed in the process of label assignment, making it less flexible.


\subsection{Rethinking Label Assignment in Dense Pedestrian Detection}

\begin{figure*}[!t]
\centering
\includegraphics[width=14cm]{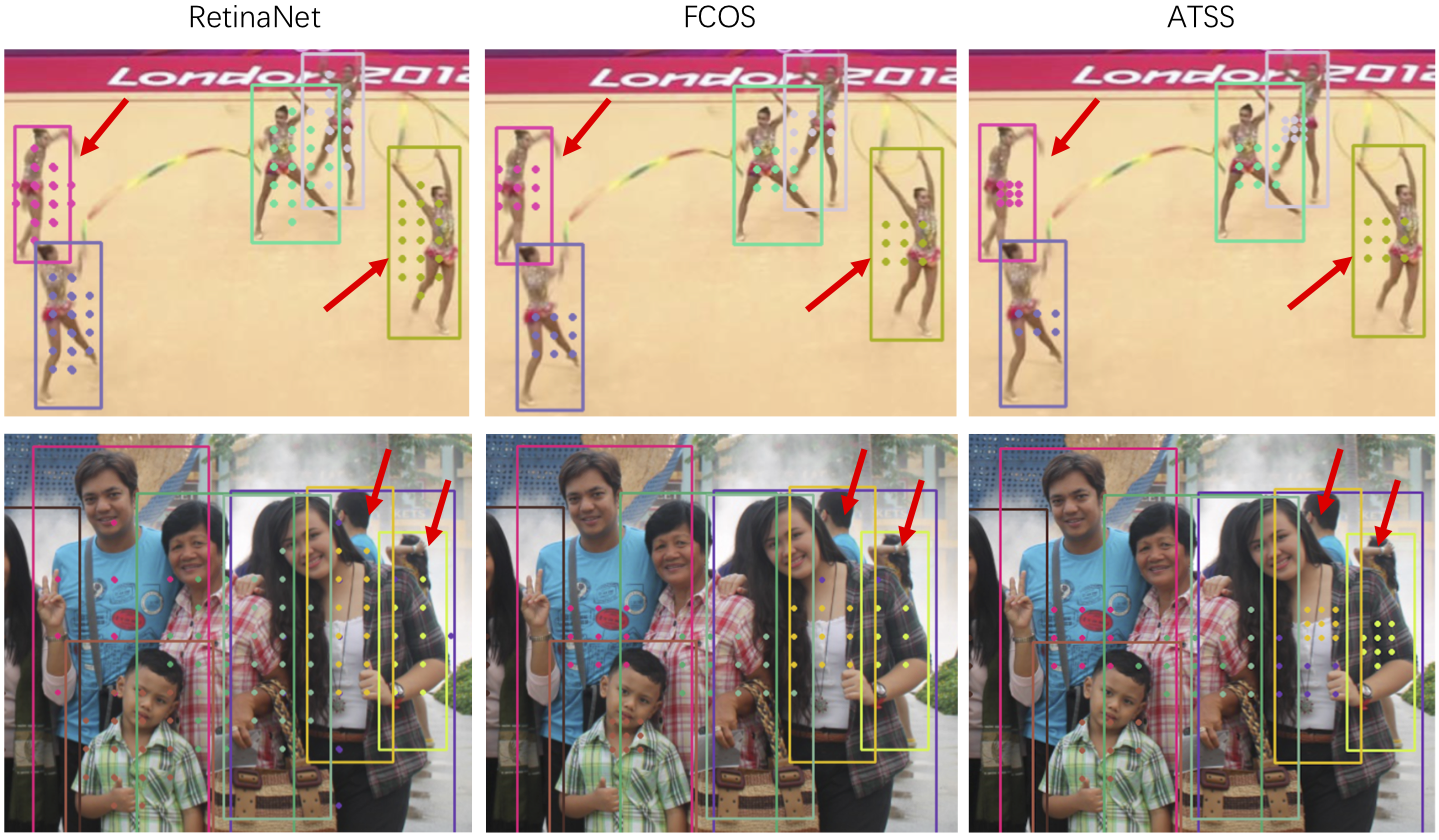}
\caption{Inappropriate label assignment in RetinaNet, FCOS and ATSS~\cite{atss}. Bounding boxes in this figure are GT annotations while dots are assigned positive anchors. \textbf{Red arrows highlight the typical inappropriate assignment}. Only FPN layers with the largest number of positive anchors are shown for better visualization.}
\label{sec3}
\end{figure*}

Label assigning strategies in RetinaNet and FCOS are based on a strong assumption that the geometric center is the most appropriate spatial location to represent objects. In statistics, this assumption may hold when we deal with objects with a large variety of categories. However, compared to general object detection, dense pedestrian detection are different in the following two aspects:

\begin{itemize}
  \item Poses between different individuals vary a lot.
  \item Human body can be heavily occluded by others.
\end{itemize}

Given these two specificities, heuristic label assigning strategies adopted in RetinaNet or FCOS are no more suitable for dense pedestrian detection. In Fig.~\ref{sec3}, we visualize some typical scenarios in which inappropriate label assignment is introduced by RetinaNet and FCOS. As can be seen in Fig.~\ref{sec3}, when the human body exhibits unusual postures (\emph{i.e.} dancing, playing sports, etc), part of positive anchors of both RetinaNet and FCOS may fall onto background regions. In these cases, the classifier will get confused and then learns improper decision boundaries. When people are very close to each other, their positive anchors will be severely intertwined where the regressor can hardly determine which target should be approached for each anchor. These examples indicate that in dense pedestrian detection, hand-crafted label assigning strategies are no more qualified.

Some recent works~\cite{atss, paa, freeanchor} have taken advantage of the outputs of models to perform label assignment, referred to as \textit{dynamic label assignment}. The success of \textit{dynamic label assignment} is based on an observation that anchors with lower loss/higher likelihood can better represent the corresponding instances. However, all these strategies need to first construct a positive candidate set of anchors for each object and then adaptively split positive and negative anchors according to the statistics from the candidate set. As the construction of the candidate set remains hand-crafted, the whole set is still near the geometric centers of their corresponding objects, which makes them \textit{partially-dynamic}. As shown in Fig.~\ref{sec3}, in ATSS, some of the positive anchors of a heavily occluded human instance, still fall into other instance's body region, which indicates that partially-dynamic label assignments are sub-optimal solutions for dense pedestrian detection.

\subsection{Loss-aware Label Assignment}

Based on previous discussion, we propose \textbf{L}oss-aware \textbf{L}abel \textbf{A}ssignment (LLA), a simple and fully-dynamic label assigning strategy for dense pedestrian detection which can be easily plugged into any anchor-based or anchor-free detectors. Without loss of generality, we extend the definition of $a_j \in A$ into an anchor box/point. Given an input image M, suppose there are $J$ anchors and $I$ GT annotations. In a single forward step, we can get score predictions $S(\theta,M) \in \mathbb{R}^{J \times N}$, where $N$ is the number of classes and $\theta$ is the learnable parameters in detector, and get bounding box predictions $B(\theta,M) \in \mathbb{R}^{J \times 4}$. Unlike previous methods which only calculate losses between each anchor and its assigned GT, LLA calculates losses between all anchor-gt pairs, getting:

\begin{equation}
\begin{split}
C^{cls} = f^{cls}(G^{cls}, S(\theta,M))\\
C^{reg} = f^{reg}(G^{loc}, B(\theta,M)),
\end{split} \label{eq1}
\end{equation}

\noindent where $C^{cls} \in \mathbb{R}^{I \times J}$ and $C^{reg} \in \mathbb{R}^{I \times J}$. $G^{cls}$ and $G^{loc}$ are ground-truth annotations for class and bounding box, respectively. $f^{cls}$ denotes binary cross entropy or Focal Loss while $f^{reg}$ can denote any of regression losses in SmoothL1~\cite{fastrcnn}, IoU and GIoU~\cite{giou} Loss. Then, a \textit{Cost Matrix} can be formulated as:

\begin{equation}
C = C^{cls} + \lambda*C^{reg},
\end{equation}

\noindent where $C \in \mathbb{R}^{I\times J}$. Given the definition of $C$, $C_{ij}$ represents the joint loss between $a_j$ and $g_i$. The smaller $C_{ij}$ is, the more possible anchor $a_j$ is assigned to GT $g_i$. Hence, we select top $K$ smallest values in each row of $C$ and consider the corresponding anchor-GT pairs \textit{matching}. 
However, in our experiments, we found that at very early training stage, LLA can hardly produce stable assignment results due to the under-fitting of the detector. To help model converge faster, we add a spatial prior -- $a_j$ can be assigned to $g_i$ only if $a_j$ (or the center of $a_j$) falls within the range of $g_i^{loc}$. Based on this prior, we introduce $C^{inbox}$

\begin{equation}
C_{ij}^{inbox}=
\begin{cases}
0,& \text{if $a_j$ in $g_i^{loc}$}\\
+\infty,& \text{if $a_j$ not in $g_i^{loc}$}.
\end{cases}
\end{equation}

In our implementation, the term $+\infty$ is replaced by a large positive value (\emph{e.g.} $10^2$). Then, the \textit{Restricted Cost Matrix} can be formulated as:

\begin{equation}
C_r = C^{cls} + \lambda*C^{reg} + C^{inbox}.
\end{equation}

After selecting the top $K$ smallest values in $C$, finally, the assignment matrix $\pi_{ij} \in \{0,1\}$ can be obtained:

\begin{equation}
\pi_{ij}=
\begin{cases}
1,& \text{if $a_j$ matches $g_i$}\\
0,& \text{if $a_j$ does not match $g_i$}.
\end{cases}
\end{equation}

Note that \textbf{if an anchor is assigned to multiple GTs simultaneously, we assign this anchor to GT with the smallest cost}. After label assigning, the detector is updates in the same as in RetinaNet and FCOS.

Compared to label assigning strategies with complex hand-crafted rules, LLA only leverages the minimal ``in box" prior. Which FPN layer should each GT assigned to is automatically determined by LLA according to the feedback of the model's outputs. Compared to partially-dynamic strategies that restrict the positive candidates near the center of each GT, LLA assigns anchors in a fully dynamic manner which helps LLA better handle the severe occlusion situations.


\begin{figure*}[!t]
\centering
\includegraphics[width=13cm]{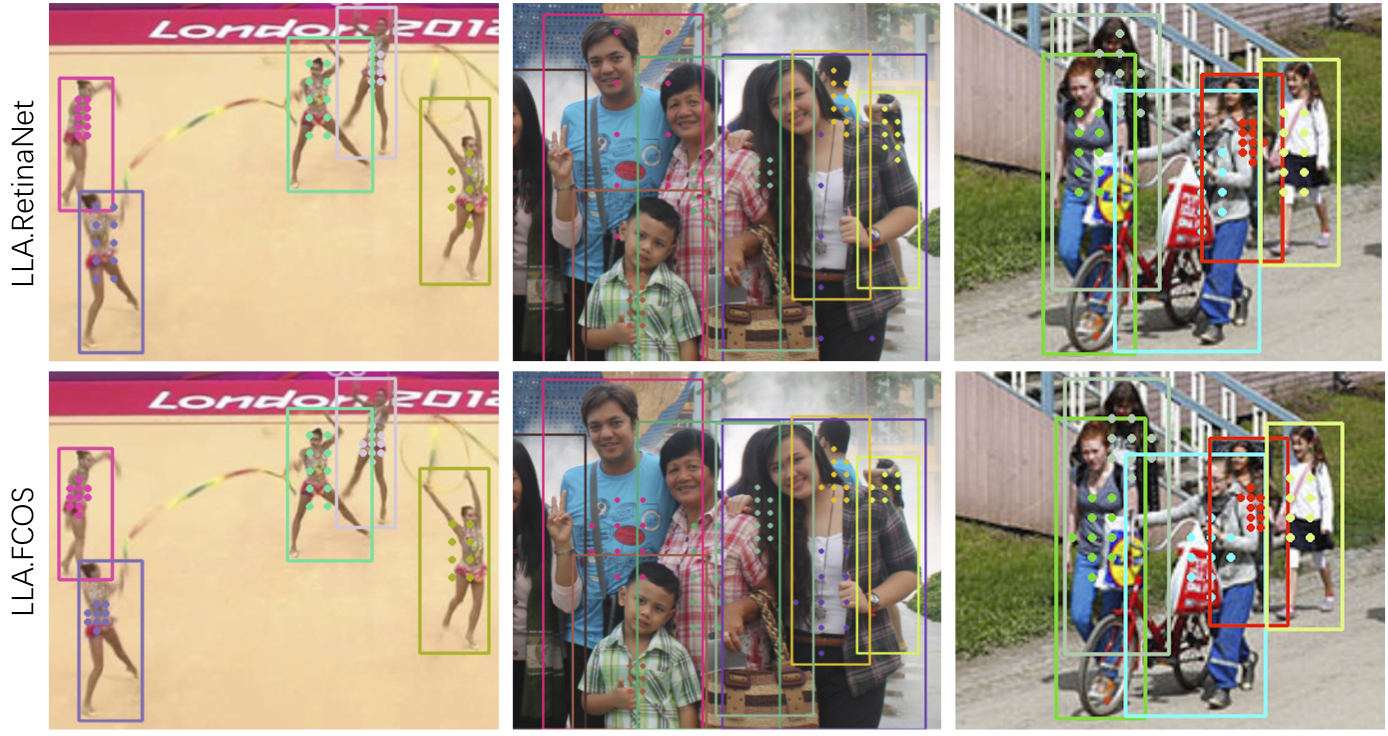}
\caption{Label assigning results of LLA. Only FPN layers with the largest number of positive anchors are shown for better visualization.}
\label{sec4}
\end{figure*}

\section{Experiments}

In this section, we carry out heavy experiments on CrowdHuman~\cite{crowdhuman} to illustrate the effectiveness of LLA. In CrowdHuman, there are 15000, 4370, and 5000 images in the training set, validation set, and testing set respectively. It provides three categories of bounding boxes annotations for each human instance: head bounding-box, human visible-body bounding-box and human full-body bounding-box. Because annotations of full-body are more crowded and challenging than visible-body, thus in this work, all of our experiments are conducted on full-body annotations. Further experimental results on CityPersons are also provided to prove LLA‘s general applicability on other datasets.

\subsection{Training Details} \label{sec4.1}

For CrowdHuman, the network structure in our experiment follows~\cite{retinanet}. We resize the input images so that the short edge is 800 pixels while the long edge is smaller than 1400 pixels. We train our model on 8 GPUs with 16 images per mini-batch. SGD with the momentum of 0.9 is adopted as our optimizer. The initial learning rate is 0.01 for both RetinaNet and FCOS and is decayed by a factor of 10 after 8th epoch and 11th epoch. The training process finishes at the end of the 12th epoch. If not specified, the default backbone in our experiments is ResNet-50~\cite{resnet}. We adopt IoU Loss for $f^{reg}$ in Eq.~\ref{eq1}. For back-propagation, the regression loss is replaced by GIoU Loss because we find such a usage of IoU/GIoU Loss can yield the best performances. For evaluation, we follow the standard Caltech~\cite{caltech} evaluation metric -- MR, which stands for the Log-Average Missing Rate over false positives per image (FPPI) ranging in $[10^{-2},100]$.

\begin{table}[!t]
\centering
\caption{Results of LLA on two detectors -- RetinaNet and FCOS. \#A=1 denotes that only one anchor on each spatial location is used. Lower is better for MR.}
\label{t41}
\vspace{5 pt}
\scalebox{0.5}{
\resizebox{\textwidth}{!}{
\begin{tabular}{c|l|ccc}
\hline
\multicolumn{1}{l|}{Backbone}                    & Method                 & MR   & AP    & Recall \\ \hline
\multirow{4}{*}{ResNet-50}                       & RetinaNet~\cite{retinanet}           & 59.13 & 81.04 & 88.25  \\ 
                                                 & RetinaNet*            & 54.00 & 80.54 & 87.33  \\ 
                                                 & RetinaNet*(\#A=1)       & 60.92 & 80.21 & 86.70  \\ 
                                                 & LLA.RetinaNet*(\#A=1) & \textbf{49.60} & \textbf{84.69} & \textbf{89.90} \\ \hline
\multicolumn{1}{c|}{\multirow{2}{*}{ResNet-50}} & FCOS~\cite{fcos}       & 54.95 & 86.36 & \textbf{94.05} \\ 
\multicolumn{1}{c|}{}                           & LLA.FCOS & \textbf{49.48} & \textbf{87.25} & 93.43 \\ \hline
\multicolumn{1}{l|}{\multirow{2}{*}{ResNet-101}} & RetinaNet             &  57.72     & 81.20  & 87.77    \\ 
\multicolumn{1}{l|}{}                            & LLA.RetinaNet*(\#A=1) &   \textbf{47.29}   & \textbf{86.25}  & \textbf{91.27} \\ \hline
\end{tabular}}}
\end{table}

\subsection{Ablation Studies}

\noindent \textbf{Effect of LLA.} We present experimental results on RetinaNet and FCOS. $\lambda$ is set to $1$ and $1.3$ while the NMS threshold is set to 0.5 and 0.6 for RetinaNet and FCOS, respectively. As seen in Table~\ref{t41}, RetinaNet achieves 59.13\% MR and 81.04\% AP on CrowdHuman. We further implement a modified version of RetinaNet, termed as RetinaNet*, in which we replace SmoothL1 Loss with recently proposed GIoU Loss to better regress instances' boundaries. To better reflect the power of LLA, we reduce the number of anchors in RetinaNet* from 9 to 1. Our modified RetinaNet* achieves worse MR but better AP and Recall. Finally, we introduce LLA into RetinaNet*. RetinaNet* with LLA improves MR and AP by a large margin -- 9.53\% on MR and 3.65\% on AP, respectively. Similar performance gain can be observed on a better backbone -- ResNet-101. For FCOS, as suggested in ATSS~\cite{atss} , we adopt a better label assigning strategy -- center sampling as our baseline. For LLA.FCOS, we remove the \textit{Centerness} branch and replace center sampling with LLA. In this case, MR and AP still get boosted by 5.47\% and 0.89\%. 

Visualizations of positive anchors assigned by LLA's are shown in Fig.~\ref{sec4}. As seen in the first column, instead of evenly distributed in each GT box, positive anchors for these dancers fall onto the foreground more compactly. In the second and third columns, positive anchors of heavily occluded human instances fall onto the visible regions (\emph{e.g.} heads, shoulders, etc.) which are far away from the geometric center of GT boxes, demonstrating the effectiveness of LLA.

\begin{table}[h]
\centering
\caption{Contributions of each component in $C_r$ on RetinaNet*. ``-'' denotes fail to converge.}
\label{component}
\vspace{5 pt}
\begin{tabular}{ccc|ccc}
\hline
$C^{cls}$ & $C^{reg}$ & $C^{inbox}$ & MR   & AP    & Recall \\ \hline
  \checkmark  &     &       & - & - & -  \\
  \checkmark  &  \checkmark   &       & 50.55 & 84.15 & 89.03  \\
  \checkmark  &  \checkmark   &    \checkmark   & 49.60 & 84.69 & 89.90  \\ \hline
\end{tabular}
\end{table}

\begin{table}[!t]
\centering
\caption{Performance of different usages of IoU/GIoU Loss. Values in bold stand for best results for MR and AP. ``-'' denotes failing to converge.}
\label{regloss}
\begin{tabular}{cc|cc}
\hline
\makecell[c]{Before \\ Assignment} & \makecell[c]{After \\ Assignment} & MR    & AP    \\ \hline
IoU               & IoU              & 49.78 & 84.48 \\
IoU               & GIoU             & \textbf{49.60} & \textbf{84.69} \\
GIoU              & IoU              & -     & -     \\
GIoU              & GIoU             & 52.52 & 80.54 \\ \hline
\end{tabular}
\end{table}

\noindent \textbf{Analysis of Each Component in Restricted Cost Matrix.} In Restricted Cost Matrix $C_r$, each term has its unique value. We start from the minimal requirement $C^{cls}$ because $C^{cls}$ helps identify the region of foreground instances. However, as seen in Table~\ref{component}, without other two terms in $C_r$, the model fails to converge. This is mainly due to that $C^{cls}$ can not help model distinguish different instances in same category by incorporating spatial information. Without term $C^{reg}$ in $C_r$, an anchor $a$ in instance $A$ can also be in topk list of instance $B$ if $A$ and $B$ are in the same category. Such a mis-assignment would lead the optimization process to wrong directions. After introducing $C^{reg}$ into $C_r$, the detector achieves 50.55\% MR which already surpasses baseline by a large margin. Based on that, $C^{inbox}$ can further improve MR by 0.95\% to help stabilize training process at early training stage.

\noindent \textbf{Different Usage of IoU/GIoU Loss.} Intersection over Union is used twice in our work: 1). Calculating regression loss between each anchor-gt pair before assignment, defined as the $C^{reg}$ term in Cost Matrix C. 2). Calculating regression loss for each assigned anchor-gt pair for back-propagation. As stated in Sec.~\ref{sec4.1}, we use IoU Loss before assignment and GIoU Loss after assignment, because we found such a usage can yield the best detection performance. Here, we present the detection performances of other different settings in Table~\ref{regloss}. Noted that our proposed LLA focuses on label assignment in object detection. The exploration of different IoU variants (e.g. DIoU~\cite{diou}, CIoU~\cite{diou}) is beyond our scope. Hence, we did not conduct further experiments.

\begin{table*}[!t]
\centering
\caption{Performance of LLA.RetinaNet under different values of $K$. Values in bold are the best results for MR and AP.}
\label{k}
\begin{tabular}{c|cccccccc}
\hline
Top $K$ & 1     & 2     & 3     & 4     & 5     & 6     & 7     & 8     \\ \hline
MR    & 53.78 & 52.19 & 51.73 & 51.25 & 50.38 & 51.38 & 50.46 & 50.06 \\
AP    & 82.72 & 83.68 & 84.14 & 84.66 & \textbf{85.64} & 84.52 & 84.42 & 84.68 \\ \hline \hline
Top $K$ & 9     & 10    & 11    & 12    & 13    & 14    & 15    & 16    \\ \hline
MR    & 49.87 & \textbf{49.60} & 49.86 & 50.03 & 50.30 & 50.16 & 50.38 & 50.54 \\
AP    & 84.88 & 84.69 & 84.30 & 84.32 & 84.30 & 84.26 & 84.00 & 83.81
\end{tabular}
\end{table*}

\noindent \textbf{Effect of $K$.} Hyper-parameter $K$ can be viewed as the number of positive anchors we want for each GT. Intuitively, too large $K$ will introduce many low-quality candidates while too small $K$ will lead to an insufficient number of candidates and then hurt the detector's accuracy. Thus we conduct heavy experiments to study the best value of $K$. We vary $K$ from 1 to 16. As shown in Table~\ref{k}, $K=10$ achieves the best MR. We also observe that $K$ is quite insensitive within a broad range which is a desired property for generalization on different datasets.


\begin{figure*}[!h]
\centering
\includegraphics[width=13cm]{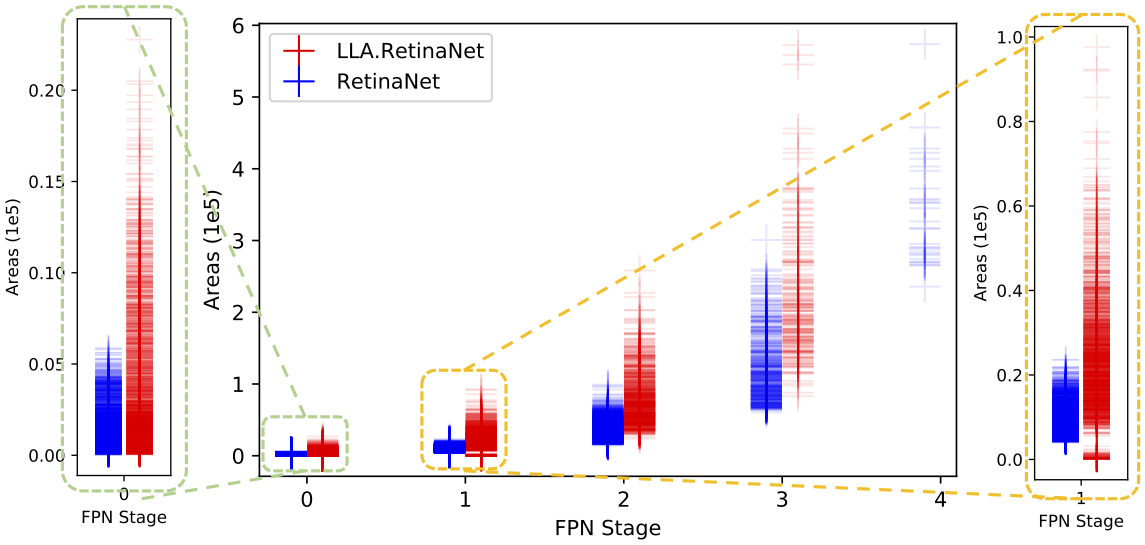}
\caption{Visualization of objects' areas and the FPN layers they are assigned to. Each light cross represents a GT box. We consider the layer which is assigned to the highest number of positive samples as the layer that GT is assigned to. ``stage 0'' denotes the highest resolution layer (\emph{i.e.} ``P3'' in FPN~\cite{fpn}).}
\label{fpn}
\end{figure*}

\subsection{Further Analysis.}

\noindent \textbf{Ambiguous Assignment Ratio.} Given a label assigning strategy, an anchor can be possibly assigned to several GTs, leading to the ambiguous assignment. For example, in RetinaNet, an anchor box may have an IoU value greater than 0.5 with multiple GTs. In this case, this anchor will be assigned to the GT with maximum IoU value. We call such kind of anchors which need further post-processing on assignment results as ambiguous anchors and define Ambiguous Assignment Ratio (AAR) as:

\begin{equation}
AAR=\frac{\#Ambiguous\ Anchors}{\#Positive\ Anchors}
\end{equation}

A lower AAR means the adopted label assigning strategy assigns positive anchors in a more deterministic way, which is a desired property in the crowd scenario. We calculate the AAR for RetinaNet and FCOS when \emph{w/} and \emph{w/o} LLA on CrowdHuman. Results shown in Table~\ref{mar} illustrate that our proposed LLA can effectively decrease the AAR and reduce the ambiguity introduced in label assignment.

\begin{table}[]
\centering
\caption{AAR evaluated on RetinaNet and FCOS when w/ and w/o LLA. As can be seen, LLA can effectively reduce AAR.}
\label{mar}
\vspace{5 pt}
\begin{tabular}{l|cc}
\hline
\multirow{2}{*}{w/ LLA} & \multicolumn{2}{c}{AAR(\%)} \\ \cline{2-3} 
                          & RetinaNet       & FCOS      \\ \hline
No                        & 7.4             & 13.2      \\
Yes                       & \textbf{6.2}             & \textbf{4.6}       \\ \hline
\end{tabular}
\end{table}

\noindent \textbf{FPN Level Allocation.} Which FPN level should each GT assigned to is crucial in label assignment. RPN and FCOS assign labels based on explicit geometric constraint (\emph{i.e.} \textit{area ranges} and \textit{regression ranges}), while RetinaNet imposes implicit scale constraint by estimating IoUs between a set of pre-defined anchors and GTs. To compare the scale constraint learned by LLA and RetinaNet, we sample 5,000 GT annotations from CrowdHuman training set, for each GT, we plot its area and corresponding FPN layer which has the largest number of positive anchors of it. The results in Fig.~\ref{fpn} show that compared to RetinaNet, LLA tends to assign objects to a more fine-grained feature layer with higher resolution, in addition, no GT box is assigned to ``P7''. Such a phenomenon is reasonable because in the crowd scenario, dense anchors are more desired to precisely assign positive anchors for those heavily occluded human instances. Thus LLA can also be termed as \textbf{occlusion-aware} which is an appealing property in many real-world applications.

\begin{figure*}[t]
\centering
\includegraphics[width=13cm]{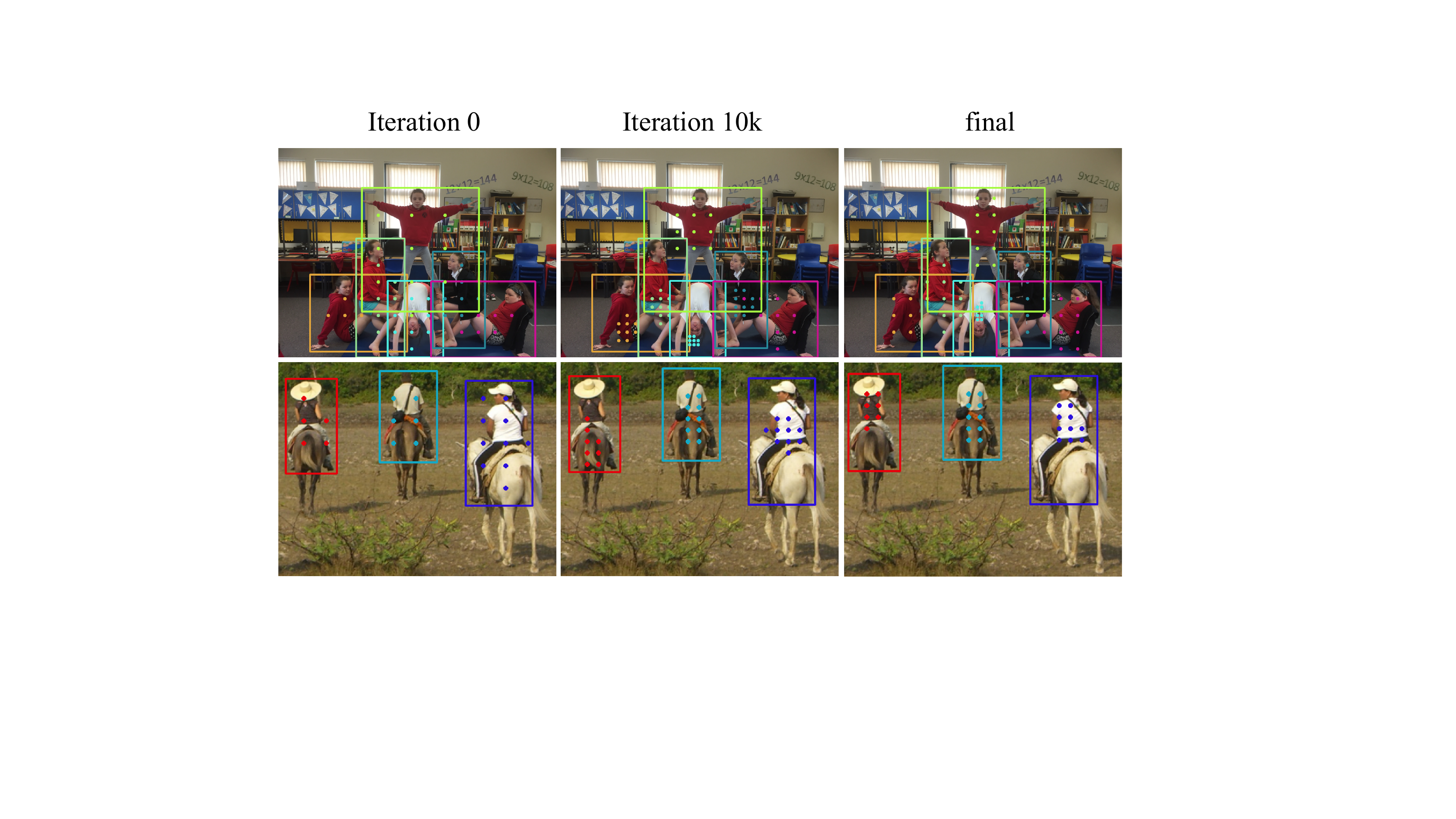}
\caption{Evolution of positive anchors during a training process.}
\label{evolution} 
\end{figure*}

\begin{table*}[!t]
\centering
\caption{Performance comparison with state-of-the-art label assigning strategies on CrowdHuman. PISA and Noisy Anchor are re-implemented based on Retinanet. For PAA, we use the official code released by its author.}
\label{sota}
\vspace{5 pt}
\begin{tabular}{lclll}
\hline
\multicolumn{1}{l|}{Method}       & \multicolumn{1}{c|}{Remarks}            & \multicolumn{1}{c}{MR}   & \multicolumn{1}{c}{AP}    & \multicolumn{1}{c}{Recall} \\ \hline \hline
\textit{Anchor-Based}             &                                         & \multicolumn{1}{c}{}      & \multicolumn{1}{c}{}      & \multicolumn{1}{c}{}       \\ \hline
\multicolumn{1}{l|}{RetinaNet~\cite{retinanet}}    & \multicolumn{1}{c|}{-}                  & \multicolumn{1}{c}{59.13} & \multicolumn{1}{c}{81.04} & \multicolumn{1}{c}{88.25}  \\
\multicolumn{1}{l|}{Noisy Anchor~\cite{noisyanchor}} & \multicolumn{1}{c|}{Based on RetinaNet} & 53.03                     & 84.01                     & 89.12                      \\
\multicolumn{1}{l|}{PISA~\cite{pisa}}         & \multicolumn{1}{c|}{Based on RetinaNet} & 52.17                     & 84.43                     & 89.68                      \\
\multicolumn{1}{l|}{FreeAnchor~\cite{freeanchor}}   & \multicolumn{1}{c|}{-}                  & 50.95       &   82.61    &     86.90   \\
\multicolumn{1}{l|}{PAA~\cite{paa}}          & \multicolumn{1}{c|}{w/ IoU Branch~\cite{paa},GN~\cite{groupnorm}}      & 50.77                     & 84.24                     & 89.60                   \\
\multicolumn{1}{l|}{LLA.RetinaNet(Ours)}  & \multicolumn{1}{c|}{-}                  & \textbf{49.60} & \textbf{84.69} & \textbf{89.90}  \\ \hline \hline
\textit{Anchor-Free}              &                                         & \multicolumn{1}{c}{}      & \multicolumn{1}{c}{}      & \multicolumn{1}{c}{}       \\ \hline
\multicolumn{1}{l|}{FCOS~\cite{fcos}}         & \multicolumn{1}{c|}{w/o Centerness}     & 71.47                          &      83.34                     &    93.76                        \\
\multicolumn{1}{l|}{}             & \multicolumn{1}{c|}{w/ Centerness}      & 54.95                     & 86.36                     & 94.05                      \\ \hline
\multicolumn{1}{l|}{ATSS~\cite{atss}}         & \multicolumn{1}{c|}{w/o Centerness}     & 56.25                     &   86.47                     & 92.88                      \\
\multicolumn{1}{l|}{}             & \multicolumn{1}{c|}{w/ Centerness}      & 49.51                     & 87.41                     & \textbf{94.19}                      \\ \hline
\multicolumn{1}{l|}{LLA.FCOS(Ours)}  & \multicolumn{1}{c|}{w/o IoU Branch}     & 49.48                     & 87.25                     & 93.43                      \\
\multicolumn{1}{l|}{}             & \multicolumn{1}{c|}{w/ IoU Branch}      & \textbf{47.90}      & \textbf{88.04}      & 93.95      \\ \hline
\end{tabular}
\end{table*}

\begin{figure*}[!t]
\centering
\includegraphics[width=17cm]{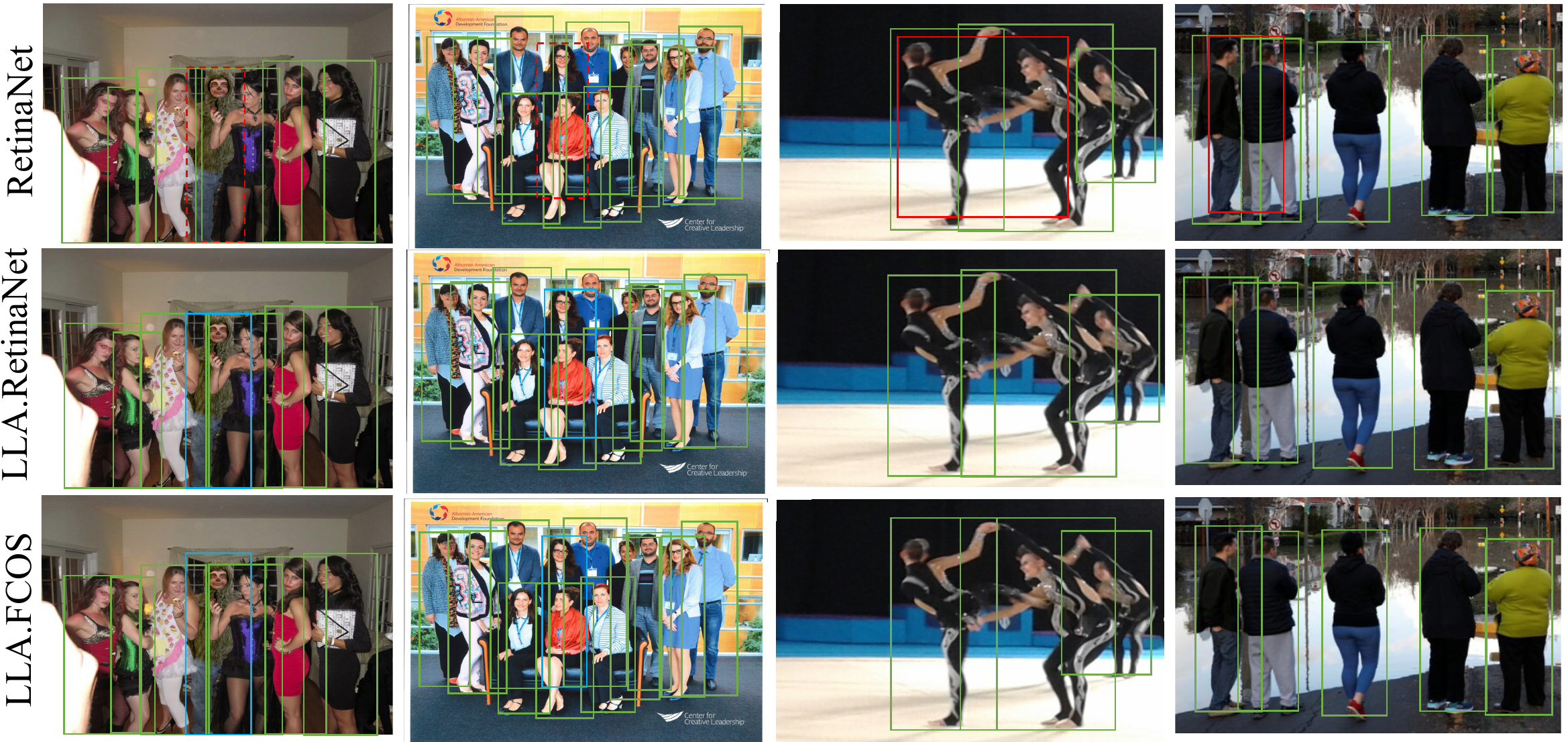}
\caption{Prediction results of RetinaNet before and after adopting LLA, as well as LLA.FCOS. Red dotted and solid lines denote miss-detections and false positives in original RetinaNet, respectively. Blue solid lines represent that the miss-detections in RetinaNet are successfully detected in LLA.RetinaNet and LLA.FCOS.}
\label{finalvis}
\end{figure*}

\noindent \textbf{Evolution of Positive Anchors.} We visualize the evolution of positive anchors during a training process in Fig.~\ref{evolution}. We use LLA.RetinaNet and only visualize the FPN layer with the largest number of positive anchors. At early training stage, due to the under-fitting of the detector, some of the assigned positive anchors may fall onto the background or on other GT's foreground. As the training process continues, positive anchors tend to migrate onto the foreground region of their corresponding GTs, demonstrating the effectiveness of LLA. Noted that although detectors like SSD~\cite{ssd}, RetinaNet~\cite{retinanet} and FCOS~\cite{fcos} tend to assign anchors close to the objects’ geometric center as their positive anchors, the success of PAA, AutoAssign~\cite{autoassign} and our proposed LLA reveals that geometric center is not the best prior. However, we do not argue that semantic center is always a better prior location for positive anchors than geometric center. Instead, we can see in Fig.~\ref{evolution} that many anchors near objects’ geometric centers are also defined positive. Utilizing geometric center or semantic center or both of them is totally adaptive and only based on model’s predictions itself. That’s why LLA can achieve SOTA performance.

\subsection{Comparison with State-of-the-art.}
Note that LLA can be compatible with recent advances in DPD, thus in this section, we only compare LLA with other state-of-the-art label assigning strategies. NMS thresholds 0.5 and 0.6 are adopted for anchor-based and anchor-free methods, respectively. As seen in Table~\ref{sota}, For the anchor-based method, our LLA built upon RetinaNet surpasses all other methods without any bells and whistles (\emph{e.g}. GroupNorm~\cite{groupnorm} and IoU Branch). For anchor-free methods, LLA is built upon FCOS. It needs to be mentioned that both FCOS and ATSS restrict positive anchors in the central region of an object, and thus they can benefit from the Centerness branch to eliminate false positives. Especially under MR -- a metric which is extremely sensitive to false positives, using Centerness can remarkably reduce MR for FCOS and ATSS. However, LLA does not acknowledge and use \textit{center prior} in crowd scenarios, under this circumstance, adopting the Centerness branch will instead hurt the detector's performance. For fair comparison, we adopt the IoU branch proposed in PAA as a replacement of Centerness. As shown in Table~\ref{sota}, LLA without IoU branch leads FCOS and ATSS \textit{w/o} Centerness by more than 5\% MR, also shows better results than ATSS \textit{w/} Centerness. After further utilizing IoU branch, LLA surpasses ATSS by a clear margin.

\subsection{Visualizing Prediction Results on CrowdHuman.}

We compare the prediction results with and without LLA in Fig.~\ref{finalvis}. The first two columns exhibit that LLA can successfully detect the mis-detected instances by original RetinaNet. The last two columns show that LLA can effectively reduce false positives. We believe such two merits in LLA mainly benefit from the better placement of positive anchors, which greatly reduces the ambiguity during the training stage.

\begin{table}[!h]
\centering
\caption{Performance comparison of different label assigning strategies on CityPersons~\cite{citypersons}.}
\label{citypersons}
\vspace{5 pt}
\begin{tabular}{lcc}
\hline
\multicolumn{1}{l|}{\multirow{2}{*}{Method}} & \multicolumn{2}{c}{MR} \\ \cline{2-3} 
\multicolumn{1}{l|}{}                        & Reasonable   & Heavy   \\ \hline \hline
\textit{Anchor-Based}                        &              &         \\ \hline
\multicolumn{1}{l|}{RetinaNet}               & 16.83        & 46.92   \\
\multicolumn{1}{l|}{FreeAnchor}              & 15.01        & 46.72   \\
\multicolumn{1}{l|}{LLA.RetinaNet(Ours)}                   & \textbf{14.34}        & \textbf{44.70}   \\ \hline \hline
\textit{Anchor-Free}                         &              &         \\ \hline
\multicolumn{1}{l|}{FCOS}                    & 15.27        & 46.82   \\
\multicolumn{1}{l|}{ATSS}                    & 13.74        & 43.86   \\
\multicolumn{1}{l|}{LLA.FCOS(Ours)}                   & \textbf{12.08}        & \textbf{43.72}
\end{tabular}
\end{table}

\subsection{Experiments on CityPersons.}

CityPersons~\cite{citypersons} is another dataset for pedestrian detection which consists of 2975 images for training, 500 and 1575 images for validation and testing. Following the evaluation protocol in CityPersons, objects whose heights are less 50 are ignored. Besides, the validation set is further divided into two subsets according to visibility -- Reasonable and Heavy Occlusion. MR on these two subsets is reported in this work. We follow the same training details as in CrowdHuman. IoU branch is adopted in Anchor Free LLA. As shown in Table~\ref{citypersons}, our anchor-based and anchor-free LLA reduce MR on the Heavy Occlusion subset by 2.22\% and 3.10\% respectively, exceeding all other existing label assigning strategies.

\section{Conclusion.}

In this work, we propose \textbf{L}oss-aware \textbf{L}abel \textbf{A}ssignmeng (LLA), an extremely simple but effective label assigning strategy for pedestrian detection in crowd scenarios. It defines positive/negative anchors based on the values of joint losses and defines anchors with smaller loss as positives. LLA does not utilize any human prior such as spatial (center/IoU constraint in ATSS/RetinaNet) and scale prior (scale constraint in FPN), making LLA fully adaptive. Experimental results on CrowdHuman and CityPersons demonstrate LLA's superiority over other label assigning strategies as well as its generalization ability on various pedestrian detection datasets.

{\small
\bibliographystyle{ieee_fullname}
\bibliography{egpaper_final}
}

\end{document}